\documentclass[letterpaper, 10 pt, journal, twoside]{IEEEtran}

\IEEEoverridecommandlockouts                              

\usepackage{graphicx}
\usepackage{caption}
\usepackage{amsmath,amssymb,enumerate}

\usepackage{amsthm}
\usepackage{mathtools}
\usepackage{algorithm, algorithmicx, algpseudocode}
\usepackage{mathrsfs}
\usepackage[mathscr]{euscript}
\usepackage{times}
\usepackage{cite} 
\usepackage{multicol}
\usepackage[caption=false,font=footnotesize]{subfig}
\usepackage{amsfonts}
\usepackage[utf8]{inputenc}
\usepackage[T1]{fontenc}
\usepackage{textcomp}
\usepackage{amsfonts}

\usepackage{multirow}
\usepackage{booktabs}
\usepackage[usenames,dvipsnames,svgnames,table, xcdraw]{xcolor}
\definecolor{buildingColor}{rgb}{0.5020, 0, 0}
\definecolor{vegetationColor}{rgb}{0.5020, 0.5020, 0}
\definecolor{carColor}{rgb}{0.2510, 0, 0.5020}
\definecolor{pedestrainColor}{rgb}{0.2510, 0.2510, 0}
\definecolor{roadColor}{rgb}{0.5020, 0.2510, 0.5020}
\definecolor{fenceColor}{rgb}{0.2510, 0.2510, 0.5020}
\definecolor{signateColor}{rgb}{0.7529, 0.5020, 0.5020}
\definecolor{sidewalkColor}{rgb}{0, 0, 0.7529}
\definecolor{poleColor}{rgb}{0.7529, 0.7529, 0.5020}

\definecolor{nwaterColor}{rgb}{0.1176, 0.5647, 0.9804}
\definecolor{nroadColor}{rgb}{0.9804, 0.9804, 0.9804}
\definecolor{nsidewalkColor}{rgb}{0.5020, 0.2510, 0.5020}
\definecolor{nterrainColor}{rgb}{0.5020, 0.5020, 0}
\definecolor{nbuildingColor}{rgb}{0.9804, 0.5020, 0}
\definecolor{nvegetationColor}{rgb}{0.4196, 0.5569, 0.1373}
\definecolor{ncarColor}{rgb}{0, 0, 0.5569}
\definecolor{npersonColor}{rgb}{0.8627, 0.0784, 0.2353}
\definecolor{nbikeColor}{rgb}{0.4667, 0.0431, 0.1255}
\definecolor{npoleColor}{rgb}{0.7529, 0.7529, 0.7529}
\definecolor{nstairColor}{rgb}{0.4824, 0.4078, 0.9333}
\definecolor{ntrafficsignColor}{rgb}{0.9804, 0.9804, 0}

\definecolor{scarColor}{rgb}{0.9608, 0.5882, 0.3922}
\definecolor{sbicycleColor}{rgb}{0.9608, 0.9020, 0.3922}
\definecolor{smotorcycleColor}{rgb}{0.5882, 0.2353, 0.1176}
\definecolor{struckColor}{rgb}{0.7059, 0.1176, 0.3137}
\definecolor{sothervehicleColor}{rgb}{1, 0.3137, 0.3922}
\definecolor{spersonColor}{rgb}{0.1176, 0.1176, 1}
\definecolor{sbicyclistColor}{rgb}{0.7843, 0.1569, 1}
\definecolor{smotorcyclistColor}{rgb}{0.3529, 0.1176, 0.5882}
\definecolor{sroadColor}{rgb}{1, 0, 1}
\definecolor{sparkingColor}{rgb}{1, 0.5882, 1}
\definecolor{ssidewalkColor}{rgb}{0.2941, 0, 0.2941}
\definecolor{sothergroundColor}{rgb}{0.2941, 0, 0.6863}
\definecolor{sbuildingColor}{rgb}{0, 0.7843, 1}
\definecolor{sfenceColor}{rgb}{0.1961, 0.4706, 1}
\definecolor{svegetationColor}{rgb}{0, 0.6863 ,0}
\definecolor{strunkColor}{rgb}{0, 0.2353, 0.5294}
\definecolor{sterrainColor}{rgb}{0.3137, 0.9412, 0.5882}
\definecolor{spoleColor}{rgb}{0.5882, 0.9412, 1}
\definecolor{strafficsignColor}{rgb}{0, 0, 1}

\definecolor{beamColor}{rgb}{0.9608, 0.5882, 0.3922}
\definecolor{boardColor}{rgb}{0.9608, 0.9020, 0.3922}
\definecolor{bookcaseColor}{rgb}{0.5882, 0.2353, 0.1176}
\definecolor{ceilingColor}{rgb}{0.7059, 0.1176, 0.3137}
\definecolor{chairColor}{rgb}{1, 0.3137, 0.3922}
\definecolor{clutterColor}{rgb}{0.1176, 0.1176, 1}
\definecolor{doorColor}{rgb}{0.7843, 0.1569, 1}
\definecolor{floorColor}{rgb}{0.3529, 0.1176, 0.5882}
\definecolor{tableColor}{rgb}{1, 0, 1}
\definecolor{wallColor}{rgb}{1, 0.5882, 1}

\newcommand\crule[3][black]{\textcolor{#1}{\rule{#2}{#3}}}

\usepackage[colorlinks=true,pdfpagemode=UseNone,citecolor=black,linkcolor=black,urlcolor=BrickRed]{hyperref}

\newtheorem{example}{Example}

\newcommand{\squeezeup}{\vspace{-3mm}}

\title{Bayesian Spatial Kernel Smoothing for Scalable Dense Semantic Mapping}

\author{Lu Gan, Ray Zhang, Jessy W. Grizzle, Ryan M. Eustice, and Maani Ghaffari
\thanks{This work was partially supported by the Toyota Research Institute (TRI), partly under award number N021515. Funding for J. Grizzle was in part provided by TRI and in part by NSF Award No.~1808051.} 
\thanks{The authors are with the University of Michigan, Ann Arbor, MI 48109, USA.
    {\tt\footnotesize ganlu@umich.edu, rzh@umich.edu, grizzle@umich.edu, eustice@umich.edu, maanigj@umich.edu}}%
}

\begin{document}

\maketitle

\begin{abstract}

This paper develops a Bayesian continuous 3D semantic occupancy map from noisy point clouds by generalizing the Bayesian kernel inference model for building occupancy maps, a binary problem, to semantic maps, a multi-class problem. The proposed method provides a unified probabilistic model for both occupancy and semantic probabilities and nicely reverts to the original occupancy mapping framework when only one occupied class exists in obtained measurements.
The Bayesian spatial kernel inference relaxes the independent grid assumption and brings smoothness and continuity to the map inference, enabling to exploit local correlations present in the environment and increasing the performance. The accompanying software uses multi-threading and vectorization, and runs at about 2 $\mathrm{Hz}$ on a laptop CPU.
Evaluations using multiple sequences of stereo camera and LiDAR datasets show that the proposed method consistently outperforms current baselines. We also present a qualitative evaluation using data collected with a bipedal robot platform on the University of Michigan - North Campus.
\end{abstract}


\begin{IEEEkeywords}
Mapping, semantic scene understanding, range sensing, RGB-D perception.
\end{IEEEkeywords}

\section{Introduction}

\IEEEPARstart{R}{obotic} mapping is the problem of inferring a representation of the robot's surroundings using noisy measurements as it navigates through an environment. This problem is traditionally solved using occupancy grid mapping techniques~\cite{moravec1985high,elfes1987sonar,hornung2013octomap}. As robotic systems move toward more challenging behaviors in more complex scenarios, such systems require richer maps so that the robot understands the significance of the scene and objects within. Hence, the integration of semantic knowledge into the map has been the focus of robotic research in recent years~\cite{nuchter2008towards,wolf2008semantic,kostavelis2015semantic,vineet2015incremental,sengupta2015semantic,yang2017semantic}.

\begin{figure}[t]
    \centering
    \subfloat{\includegraphics[width=0.75\columnwidth]{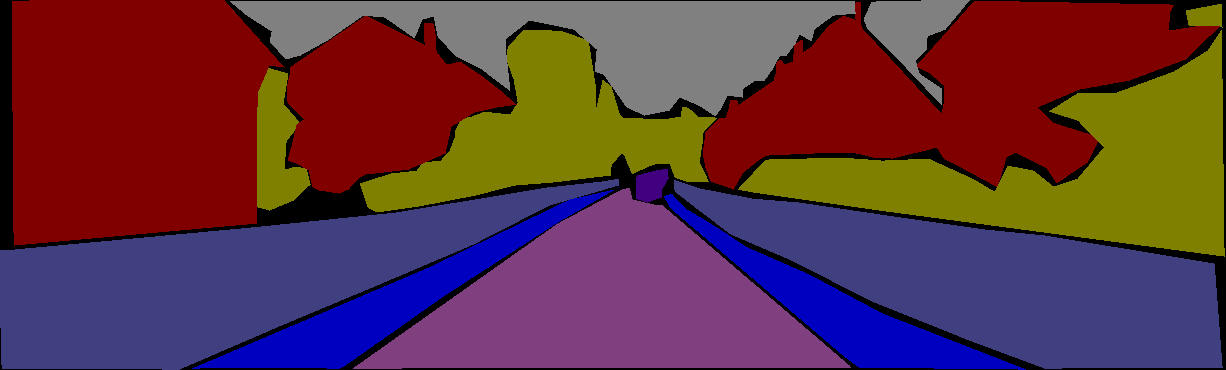}}\\ \vspace{-3mm}
    \subfloat{\includegraphics[width=0.75\columnwidth]{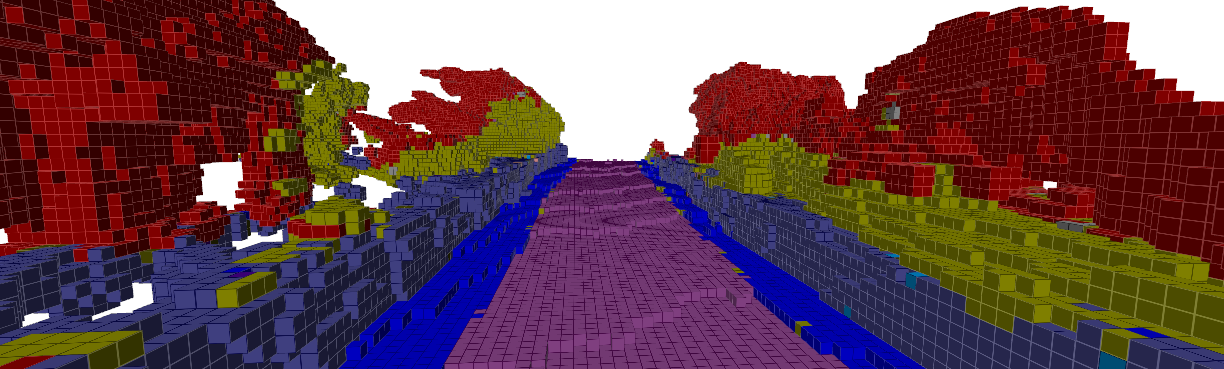}}\\ \vspace{-3mm}
    \subfloat{\includegraphics[width=0.75\columnwidth]{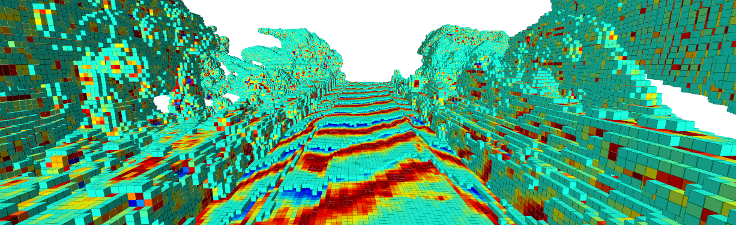}} 
    \caption{Qualitative results on KITTI odometry sequence 05 dataset~\cite{kundu2014joint}. From top to bottom the figures show the 2D ground truth image, 3D semantic map, and variance map.}
    \label{fig:kitti_05_proj_first}
\end{figure}

A semantic occupancy map as shown in Fig.~\ref{fig:kitti_05_proj_first}, besides possessing properties similar to an occupancy grid map,  maintains for each cell a set of probabilities of semantic classes. These probabilities are often updated using a Bayes filter~\cite{stuckler2012semantic,yang2017semantic}, and then Conditional Random Fields (CRF) or Markov Random Fields (MRF) are subsequently applied to mitigate discontinuities and inconsistencies in the semantic map~\cite{kim20133d, vineet2015incremental, sengupta2015semantic, yang2017semantic, zhao2016building}. In principle, CRF models encourage label consistency among neighboring grids in super-voxels~\cite{sengupta2015semantic} or 2D superpixels~\cite{zhao2016building, yang2017semantic}. However, CRF optimization is only 
applied as a post-processing step, and therefore, it is unable to predict semantics of partially observed regions in the map.

Occupancy grid maps assume the grids are statistically independent. However, a series of investigations on continuous occupancy mapping shows that taking local spatial correlations into account increases mapping performance~\cite{o2012gaussian,gpmap,jadidi2014exploration,wang2016fast,ramos2016hilbert,mjadidi-2018a,doherty2017bayesian,doherty2019learning}. Building on a similar idea, continuous semantic maps~\cite{ghaffari2017gaussian,gan2017sparse} can deal with sparse sensor measurements by inferring semantics of partially observed regions from neighboring measurements. Recent work on Bayesian generalized kernel inference for occupancy map prediction (BGKOctoMap) proposed in~\cite{doherty2019learning} uses a kernel inference approach to generalize the counting sensor model\cite{hahnel2003map} to continuous maps while maintaining the scalability of the method. 

In this paper, we extend BGKOctoMap~\cite{doherty2019learning} to continuous semantic mapping where the inference reverts to the original framework when only one occupied class exists.
In particular, the contributions of this work are 1) we develop a continuous statistical model for semantic occupancy mapping which models occupancy and semantic probabilities in a unified framework and queries can be made at any resolution; 2) we provide an open-source implementation of the proposed method. The current implementation exploits multi-threading and vectorization and can be run at about 2 $\mathrm{Hz}$ using a laptop CPU; 3) we present extensive experiments using both stereo camera and LiDAR data. The evaluations show that the proposed method consistently outperforms state-of-the-art systems.





Related work is given in Section~\ref{sec:related_work}. Section~\ref{sec:csm} presents preliminaries and semantic counting sensor model. Section~\ref{sec:bgki} describes an extension to continuous mapping. Experimental results are presented in Section~\ref{sec:exp}. Limitations of this work are discussed in Section~\ref{sec:discussions} and Section~\ref{sec:conclusion} concludes the paper.

\section{Related Work}
\label{sec:related_work}


\textbf{Discrete 3D Semantic Mapping.} Early semantic mapping work uses traditional pixel-wise image segmentation methods and directly transfers image labels from 2D to 3D. Labels from multiple images are fused in 3D without any further 3D optimization~\cite{he2013nonparametric, sengupta2013urban, stuckler2012semantic}. He et al.~\cite{he2013nonparametric} build a semantic octomap by using an MRF for image segmentation and selecting the most frequent label of the 3D points inside each grid.
Sengupta et al.~\cite{sengupta2013urban} build a semantic volumetric map by adopting a CRF for 2D semantic segmentation and assigning labels by a voting scheme. Stückler et al.~\cite{stuckler2012semantic} use random decision forests to segment object classes in images and fuse soft labels in a voxel-based 3D map using a Bayesian update. While these methods are similar to our semantic counting sensor model in a way that the maximum of semantic labels in a 3D element is picked in label fusion, the latter is a closed-form Bayesian inference which outputs the mean and variance of the posterior. 

To deal with noisy 2D predictions, 3D CRF optimization has been introduced as a refinement technique and it is widely applied in 3D semantic mapping~\cite{kim20133d,valentin2013mesh,vineet2015incremental}. In~\cite{kundu2014joint,sengupta2015semantic,zhao2016building}, a higher-order dense CRF model is used to further optimize the semantic predictions for 3D elements. Basic CRF models encourage label consistency for adjacent 3D elements, while higher-order dense CRFs can model long-range relationships within a region, such as grids in super-voxels~\cite{sengupta2015semantic} or grids corresponding to 2D superpixels~\cite{zhao2016building}, and further improve the mapping performance. 
More recent work uses deep Convolutional Neural Networks (CNNs) for 2D image segmentation, and follows the same framework for building 3D semantic maps~\cite{mccormac2017semanticfusion,yang2017semantic}. However, CRF optimization post-processes the inferred occupied grids, which does not change the principle of discrete semantic map inference. In~\cite{chen2019iros}, a semantic Simultaneous Localization and Mapping (SLAM) system, SuMa++, builds a surfel-based semantic map using SemanticKITTI dataset~\cite{behley2019iccv} as its byproduct. However, surfel-based maps do not model occupied or free space, thus are not used for robot navigation.

\textbf{Continuous Mapping.} Gaussian Process Occupancy Map (GPOM)~\cite{o2012gaussian} takes into account the correlation between map points and treats the map inference as a binary classification at an arbitrary resolution.
Hilbert maps~\cite{ramos2016hilbert} are more scalable and can be updated in linear time where a logistic regression classifier is trained online through stochastic gradient descent.
GPOM has been extended from binary to multi-class case in~\cite{ghaffari2017gaussian}. However, the complexity of the model grows with the number of data points, resulting in $\mathcal{O}(n^3)$ cost without approximation. 
The cost also grows with the number of semantic classes as a one-vs.-rest approach is used to build the multi-class classifier. 
Similarly, Hilbert maps can also be extended to the multi-class maps using a multinomial model. However, as discussed in~\cite{doherty2019learning}, 
the logistic regression classifier used by Hilbert map-based approaches does not provide associated uncertainties in probability estimates.

\textbf{Bayesian Kernel Inference.} Bayesian Kernel Inference (BKI) was introduced in~\cite{vega2014nonparametric} as an approximation to Gaussian processes that requires only $\mathcal{O}(\log N)$ computations instead of $\mathcal{O}(N^3)$, where $N$ is the number of training points. It generalizes local kernel estimation to the context of Bayesian inference for the exponential family of distributions. Instead of approximating inference on the model, the approximation is made at the stage of model selection. Assuming latent training parameters are conditionally independent given the target parameters, exact inference on this model is possible for any likelihood function from the exponential family. In~\cite{peretroukhin2016probe}, BKI is successfully applied to a visual odometry problem for modeling sensor uncertainty. 
In~\cite{richter2018bayesian}, BKI has been used on a Bernoulli-distributed random event with Beta-distributed prior to model collision in safe high-speed navigation problems and could achieve safe behavior in a novel environment with no relevant training data. BKI was first used in the context of mapping problems in~\cite{doherty2017bayesian,doherty2019learning}, to generalize the discrete counting sensor model~\cite{hahnel2003map} to continuous occupancy mapping.
Later, the applications of BKI in elevation regression and traversability classification are explored 
in~\cite{shan2018bayesian}. Following the same idea, we apply BKI in our semantic counting sensor model and generalize it to continuous semantic mapping. In particular, we use BKI on a Categorical likelihood with a Dirichlet distribution as its conjugate prior.


\section{Preliminaries and Semantic Counting Sensor Model}
\label{sec:csm}
The counting sensor model describes occupancy probability via a Bernoulli likelihood function. It counts for each grid how often a beam has ended in that grid and how often a beam has passed through it. This model has comparable performance to Bayesian updates in occupancy grid mapping~\cite{hahnel2005mapping}. The semantic counting sensor model is its natural generalization from occupancy (binary) mapping to semantic (multi-class) mapping.

Let $\mathcal{K} = \{1, 2, ..., K\}$ be the set of semantic class labels, i.e., $K$ categories, and $\mathcal{X} \subset \mathbb{R}^3$ be the map spatial support. For any map point $x_i \in \mathcal{X}$, we have a one-hot-encoded measurement tuple $y_i = (y_i^1, ..., y_i^K)$, where $y_i^k\ge 0$ and $\sum_{k=1}^K y_i^k = 1$. In practice, $y_i$ is the output of a max function computed using the output of a deep network for multi-class classification. The training set (data) can be defined as $\mathcal{D} := \{(x_i, y_i)\}_{i=1}^N$.

Assuming map cells are indexed by $j \in \mathbb{Z}^+$, the $j$th map cell can take on one of $K$ possible categories with the probability of each category separately specified as $\theta_j = (\theta^1_j, ..., \theta^K_j)$, where $\sum_{k=1}^{K} \theta_j^k = 1$. The $j$th map cell with semantic probability $\theta_j$ is described by a Categorical distribution as:
\begin{equation}
    p(y_i | \theta_j) = \prod_{k=1}^{K} \left ( {\theta_j^k} \right )^{y_i^k}.
\label{eq:categorical}
\end{equation}
In semantic mapping, we seek the posterior over $\theta_j$; $p(\theta_j | \mathcal{D})$.

For incremental Bayesian inference, we adopt a Dirichlet prior distribution over $\theta_j$, given by $Dir(K, \alpha_0)$, as the conjugate prior of the Categorical likelihood, where \mbox{$\alpha_0 = (\alpha_0^1, ..., \alpha_0^K)$}, $\alpha_0^k \in \mathbb{R}^{+}$ are concentration parameters (hyperparameters). Applying Bayes' rule, the posterior is given by $Dir(K, \alpha_j)$, \mbox{$\alpha_j = (\alpha_j^1, ..., \alpha_j^K)$}, where $\alpha_j^k$ is
\begin{equation}
\label{eq:alpha}
    \alpha_j^k := \alpha_0^k + \sum_{i, \text{ $x_i$ in cell } j} y_i^k.
\end{equation}

Because $\alpha_j^k$ counts the number of measurements which falls into the $j$th cell and indicate the $k$th category, we call this model the Semantic Counting Sensor Model \mbox{(S-CSM)}. Given concentration parameters $\alpha_j$, the mode of $\theta_j$ has the following closed form, which is also the maximum-a-posteriori estimate of $\theta_j$:
\begin{equation}
\label{eq:mode}
    \hat{\theta}_j^k = \frac{\alpha_j^k-1}{\sum_{k=1}^K \alpha_j^k - K}\ \text{and}\ \alpha_j^k > 1.
\end{equation}
We also have the closed-form expected value and variance of $\theta_j$ as follows:
\begin{equation}
\label{eq:variance}
    \mathbb{E}[\theta_j^k] = \frac{\alpha_j^k}{\sum_{k=1}^K \alpha_j^k}\ \text{and}\
    \mathbb{V}[\theta_j^k] = \frac{\frac{\alpha_j^k}{\sum_{k=1}^K \alpha_j^k}(1-\frac{\alpha_j^k}{\sum_{k=1}^K \alpha_j^k})}{\sum_{k=1}^K \alpha_j^k+1}.
\end{equation}
We use~\eqref{eq:alpha} to calculate the parameters of the posterior Dirichlet distribution for cell $j$ and given the posterior parameter $\alpha_j$, the statistics of 
cell $j$ can be computed by~\eqref{eq:mode}~and~\eqref{eq:variance}.

For free-class measurements, we use free-space points linearly interpolated along each sensor beam.
We note that in the particular case when $K=1$ represents the free-space class and $K=2$ represents the occupied class, the semantic counting sensor model nicely reverts to the original counting sensor model.

However, the semantic counting sensor model inherits the traditional occupancy grid mapping limitations because the posterior parameters for each cell are only correlated with measurements that directly fall into or pass through that cell. To mitigate this shortcoming, we use BKI to convert the discrete semantic counting sensor model to a continuous model by taking into account local correlations in the map.

\section{Continuous Semantic Mapping via Bayesian Kernel Inference}
\label{sec:bgki}

Bayesian kernel inference, as introduced by Vega-Brown et al.~\cite{vega2014nonparametric}, relates the extended likelihood $p(y_i | \theta_\ast, x_i, x_\ast)$ and the likelihood $p(y_i | \theta_i)$ by a smoothness constraint, where $\theta_\ast$ is the value of the latent variable for the query point $x_\ast$. In this framework, the maximum entropy distribution $g$, satisfying $D_{\text{KL}}(g \Vert f)$, has the form $g(y) \propto f(y)^{k(x_\ast, x)}$, where $D_{\text{KL}}(\cdot \Vert \cdot)$ is the Kullback-Leibler Divergence (KLD), and $k(\cdot, \cdot)$ is a kernel function. Let $g$ be the extended likelihood and $f$ the likelihood, we define a smooth distribution over semantics as having bounded KLD between the two distributions. Given a kernel function operating on 3D spatial inputs $k: \mathcal{X} \times \mathcal{X} \to [0, 1]$, we have
\begin{equation}{}
\prod_{i=1}^N p(y_i | \theta_\ast, x_i, x_\ast) \propto \prod_{i=1}^N p(y_i | \theta_\ast)^{k(x_\ast, x_i)}.
\label{eq:kldivergence}
\end{equation}

Using Bayes' rule, we can write
\begin{equation}
    p(\theta_\ast | x_\ast, \mathcal{D}) \propto p(\mathcal{D} | \theta_
    \ast, x_\ast) p(\theta_\ast | x_\ast),
\label{eq:posterior}
\end{equation}
and by substituting \eqref{eq:kldivergence} into \eqref{eq:posterior}, we have:
\begin{equation}
        p(\theta_\ast | x_\ast, \mathcal{D}) \propto 
        \left[ \prod_{i=1}^N p(y_i | \theta_\ast)^{k(x_\ast, x_i)} \right] p(\theta_\ast | x_\ast).
\end{equation}

We adopt the Categorical likelihood and place a prior distribution $Dir(K, \alpha_0)$ over $\theta_\ast$. Subsequently, \eqref{eq:posterior} becomes:
\begin{align}
    \nonumber p(\theta_\ast | x_\ast, \mathcal{D})  & \propto 
    \left[ \prod_{i=1}^N \left[ \prod_{k=1}^{K} \left ( {\theta_\ast^k} \right )^{y_i^k} \right]^{k(x_\ast, x_i)} \right]
    \prod_{k=1}^K \left ( {\theta_\ast^k} \right )^ {\alpha_0^k - 1} \\
    & = \prod_{k=1}^K \left ( {\theta_\ast^k}\right )^{\alpha_0^k + \sum_{i=1}^N y_i^k k(x_\ast, x_i) - 1},
\end{align}
which is proportional to the posterior $Dir(K, \alpha_\ast)$ where \mbox{$\alpha_\ast = (\alpha_\ast^1, ..., \alpha_\ast^K)$} is defined as
\begin{equation}
\label{eq:alpha_kernel}
    \alpha_\ast^k := \alpha_0^k + \sum_{i=1}^N k(x_\ast, x_i) y_i^k .
\end{equation}
The mode, mean, and variance for the continuous model can be computed exactly as given in \eqref{eq:mode} and \eqref{eq:variance}.

Compared with \eqref{eq:alpha}, \eqref{eq:alpha_kernel} not only considers measurements which fall into a cell but also adjacent measurements with a weighting coefficient defined by the kernel function, i.e., the distance to the query point. We note that the kernel neither needs to be positive-definite nor symmetric. To reduce the computational complexity, we choose the sparse kernel~\cite{melkumyan2009sparse} as
\begin{align}
\label{eq:kernel}
\nonumber &k(x,x') = \\
    &\begin{cases}
      \sigma_0 \left[ \frac{1}{3} \left( 2+\cos{(2\pi\frac{d}{l})} (1-\frac{d}{l})+\frac{1}{2\pi} \sin{(2\pi\frac{d}{l})}\right) \right] & \text{if}\ d < l \\
      0 & \text{if}\ d \geq l
    \end{cases}
\end{align}
where $d = \lVert x - x' \rVert$, $l > 0$ is the length-scale, and $\sigma_0$ is kernel scale parameter (signal variance). 

The derived continuous semantic model can deal with sparse and noisy sensor measurements better and allows for queries at an arbitrary resolution. 
In the context of semantic occupancy mapping, the query points are chosen to be the grid centroids. Thus, \eqref{eq:alpha_kernel} can be used to recursively update the posterior parameters for each grid.
We use a block to contain a number of grids according to the block depth, where each block is an octree of grids. For every block of test data, the corresponding training data is comprised of all portions of the new measurements that pass through the block's extended block~\cite{wang2016fast}, which is defined as the set of neighboring blocks with faces adjacent to the block containing the test data of interest. 

\begin{figure}[t!]
    \centering
    \subfloat[]{\includegraphics[width=0.35\columnwidth]{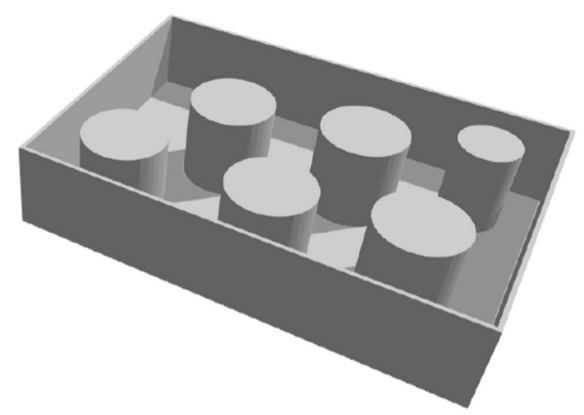}}
    \subfloat[]{\includegraphics[width=0.35\columnwidth]{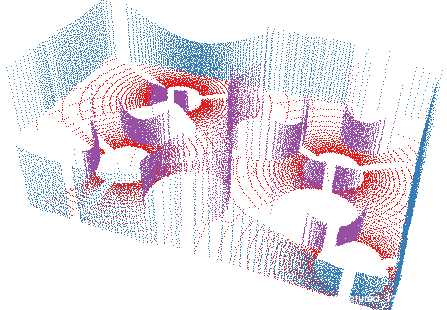}}\\ \vspace{-3mm}
    \subfloat[]{\includegraphics[width=0.35\columnwidth]{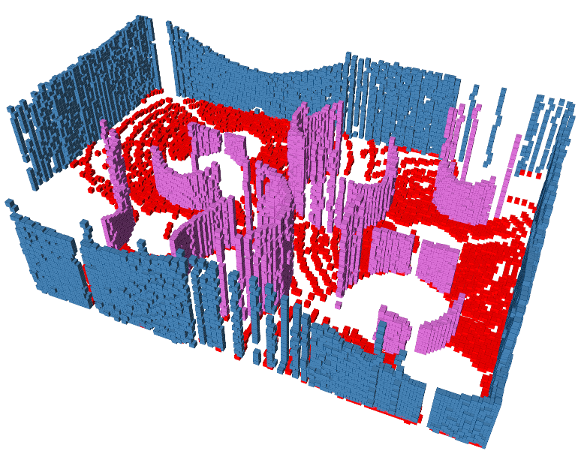}}
    \subfloat[]{\includegraphics[width=0.35\columnwidth]{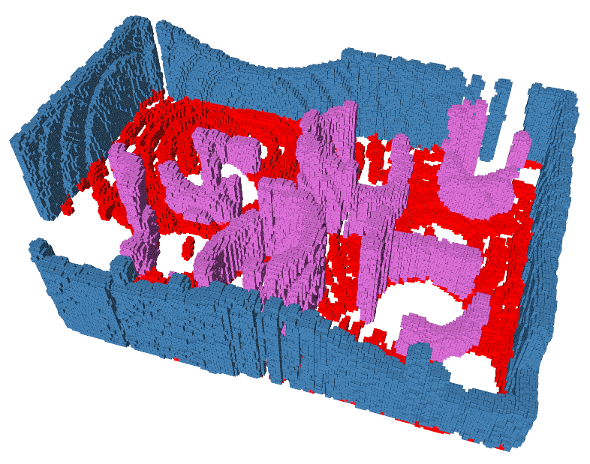}}\\ \vspace{-3mm}
    \subfloat[]{\includegraphics[width=0.35\columnwidth]{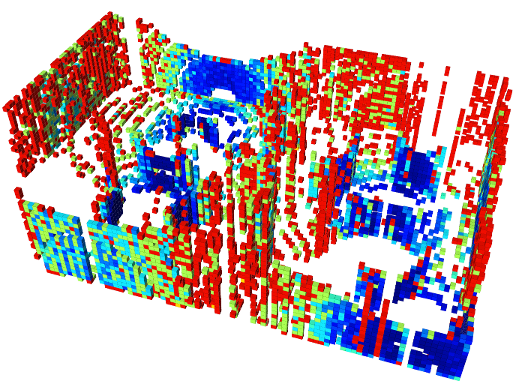}}\label{fig:variance_csm}
    \subfloat[]{\includegraphics[width=0.35\columnwidth]{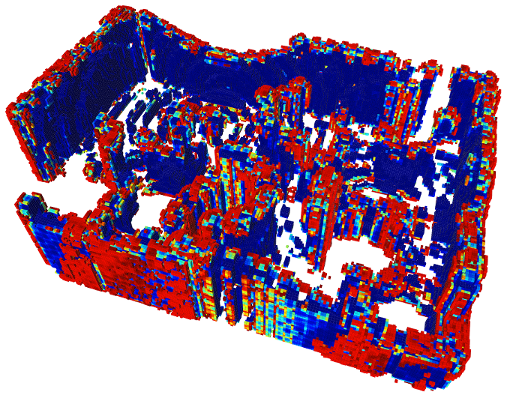}}\label{fig:vairance_bgk}
    \caption{3D toy example on a simulated dataset.  (a) Environment model in Gazebo. (b) Annotated point cloud raw data. (c) Semantic map of S-CSM. (d) Semantic map of S-BKI. (e) Variance map of S-CSM. (f) Variance map of S-BKI. Variance maps of two models (shown using the jet colormap) provide useful information for robotic navigation and exploration~\cite{mghaffari2019sampling}. We found that Bayesian kernel inference decreases the variance of the wall by considering neighboring measurements. There are some artifacts, however, on the periphery of the wall where the variance is relatively high.}
    \label{fig:exp}
\end{figure}

\begin{example}[Three-dimensional Toy Example] 
Figure~\ref{fig:exp} illustrates a three-dimensional toy example of the continuous semantic mapping via Bayesian kernel inference using a simulated dataset made in Gazebo, with annotated semantic labels. The simulated dataset has dimensions $10.0 \times 7.0 \times 2.0 m$. We manually annotate the raw data into three semantic classes: ground, wall, and cylindrical obstacles. Semantic occupancy maps with resolution 0.05 $\mathrm{m}$ for both S-CSM and Semantic Bayesian Kernel Inference (S-BKI) models are built using the annotated point clouds as sensor measurements. The figure shows that S-CSM can reconstruct the 3D environment with correct semantic information but has a limited predictive capability where sensor coverage is sparse. The S-BKI map can interpolate the gaps in the walls due to the continuity and smoothness of Bayesian kernel inference.
\end{example}

\begin{figure*}[t]
    \centering
    \vspace{-2mm}
    \subfloat{\includegraphics[width=0.5\columnwidth]{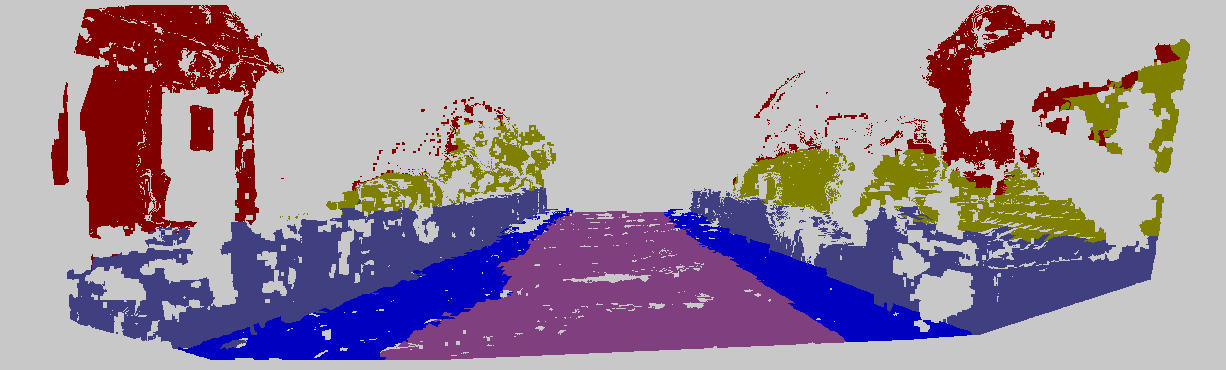}}~
    \subfloat{\includegraphics[width=0.5\columnwidth]{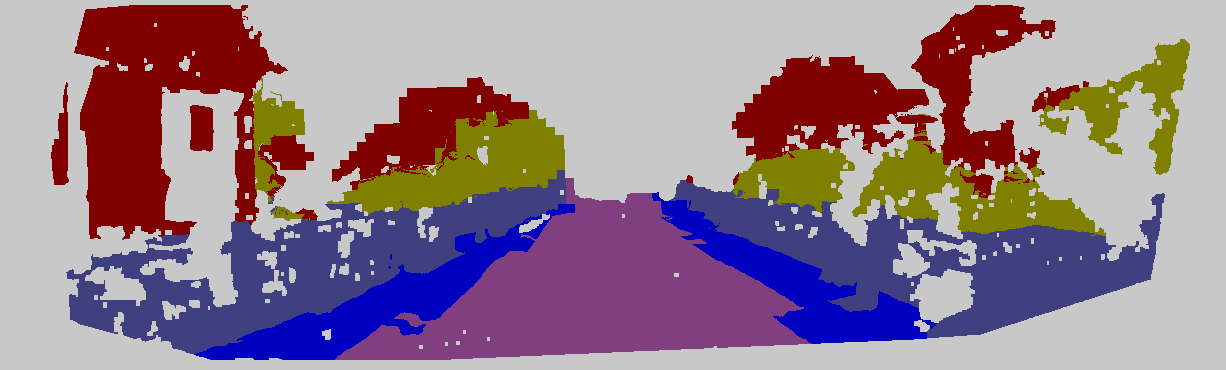}}~
    \subfloat{\includegraphics[width=0.5\columnwidth]{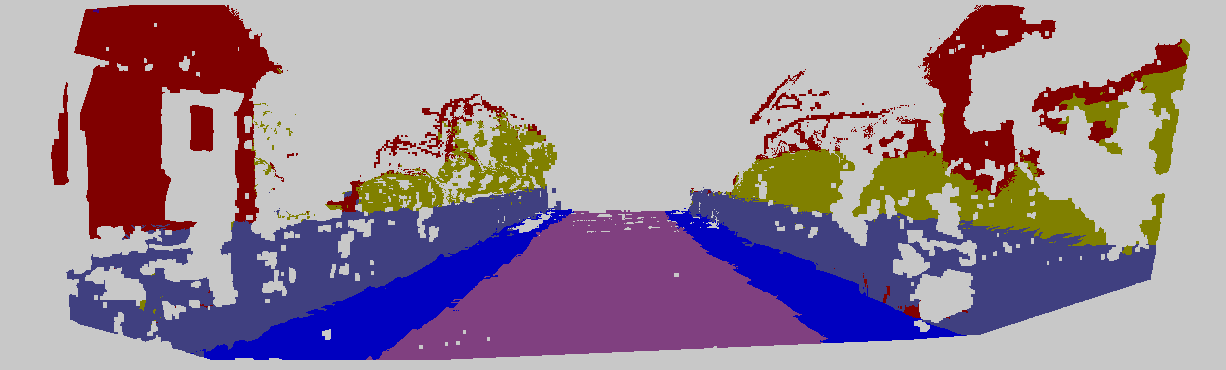}}~
    \subfloat{\includegraphics[width=0.5\columnwidth]{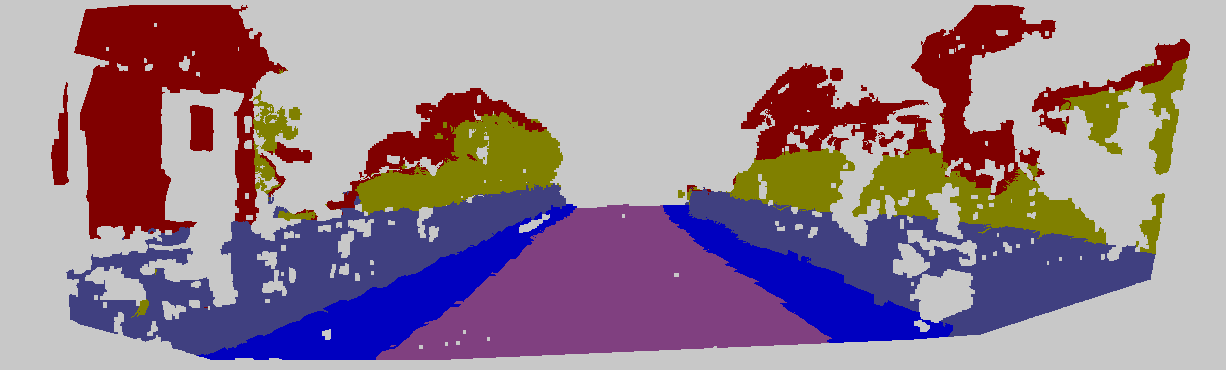}} 
\caption{Qualitative results on KITTI odometry sequence 05 dataset~\cite{kundu2014joint}. From left to right the figures show 2D projected images from Yang et al.~\cite{yang2017semantic}, BGKOctoMap-CRF, S-CSM and S-BKI, respectively. The projected image from Yang's semantic map contains more gaps than other maps, compared with the ground truth image where the road, buildings, and vegetation are continuous and dense, while the projected image of S-BKI has the least holes in those regions, which resembles the ground truth better. BGKOctoMap-CRF outperforms Yang's method, in spite of the misclassification of the sidewalk to road.}
    \label{fig:kitti_05_proj}
\end{figure*}

\section{Experimental Results}
\label{sec:exp}

We now present experiments for evaluating semantic segmentation accuracy, occupancy prediction accuracy, and the impact of parameters using multiple real datasets. We also compare the proposed methods with state-of-the-art systems and present a qualitative evaluation using data collected with a bipedal robot.
C++ implementations of the proposed methods are available open source~\footnote{\href{https://github.com/ganlumomo/BKISemanticMapping}{https://github.com/ganlumomo/BKISemanticMapping}}, and make use of the Learning-Aided 3D Mapping Library~\cite{doherty2019learning}, the Robot Operating System (ROS)~\cite{quigley2009ros}, and Point Cloud Library (PCL)~\cite{rusu20113d}. The parameters in Table~\ref{table:parameters} were manually tuned but remained fixed throughout all experiments. For baselines, we used the available open-source implementations without any modification. All experiments are conducted on an Intel i7 processor with 8 cores and 32 GB RAM.

\begin{table}[t]
\centering
\caption{Kernel and Dirichlet prior hyperparameters for all experiments.}
\label{table:parameters}
\begin{tabular}{lll}
\toprule
 Hyperparameter & Description & Value \\
 \midrule
 $l$ & Kernel length-scale  & 0.3 $\mathrm{m}$\\
 \midrule
 $\sigma_0$ & Kernel scale & 0.1\\
 \midrule
 $\alpha_0^k$ & Dirichlet prior & 0.001 \\
 \bottomrule
\end{tabular}
\end{table}

\subsection{KITTI Dataset}
KITTI dataset with semantically labeled images contains 40 test images from sequence 05~\cite{kundu2014joint}, and 25 test images from sequence 15~\cite{sengupta2013urban} in KITTI odometry dataset. We qualitatively and quantitatively compare the mapping performance of our methods with the state-of-the-art CRF-based semantic mapping system proposed by Yang et al.~\cite{yang2017semantic}. However, Yang's method only predicts semantic labels on occupied voxels using a discrete occupancy grid mapping algorithm. For a fair comparison with respect to the occupancy model, we implement another baseline, BGKOctoMap-CRF~\footnote{\href{https://github.com/zeroAska/BGKOctoMap-CRF}{https://github.com/zeroAska/BGKOctoMap-CRF}}, by replacing Yang's discrete occupancy grid map with the continuous BGKOctoMap, and then applying the same hierarchical CRF model to refine the voxel labels.

We adopt the same data pre-processing methods as used by Yang et al.~\cite{yang2017semantic}. We use ELAS~\cite{geiger2010efficient} to generate depth maps from stereo image pairs, ORB-SLAM~\cite{mur2015orb} to estimate 6DoF camera poses, and the deep network dilated CNN~\cite{yu2015multi} for prior semantic label predictions. The superpixels used in Yang's CRF module and BGKOctoMap-CRF are generated by the SLIC algorithm~\cite{achanta2012slic}. The common parameters for occupancy mapping in all methods are set according to Yang's work: the resolution of 0.1 $\mathrm{m}$, free and occupied thresholds as 0.47 and 0.6, respectively.

\subsubsection{Qualitative Results}
The 3D view of the semantic map built by S-BKI model is given in Fig.~\ref{fig:kitti_05_proj_first}. Our approach is able to recognize and reconstruct general objects such as road, sidewalk, building, fence and vegetation. We also show the same view of the corresponding variance map of S-BKI in Fig.~\ref{fig:kitti_05_proj_first}. Most of the grids on the surface have relatively low variance (cyan); the middle grids have the lowest variance (blue) where the sensor measurements are dense, while the grids on the margins of sensor scans show relatively high variance (red) where the sensor measurements are sparse. It can also be noticed that the uneven parts of the road in the semantic map have high variance, which might be caused by the discontinuity of the estimated camera poses. 

We also found that a small portion of grids of the fence on the left side are misclassified as vegetation, where the corresponding variance is high. This nice property enables us to reject misclassified grids by setting a variance threshold. If the variance is too high, we can regard the state of the grid as unknown and thus build safer semantic maps for robot navigation. To compare the mapping performance, we project semantic maps onto 2D left camera views and compare with 2D ground truth images as shown in Fig.~\ref{fig:kitti_05_proj}.

\subsubsection{Quantitative Results}
We follow the evaluation method given in~\cite{yang2017semantic} by projecting 3D semantic map onto the 2D left image plane, ignoring voxels that are too far from the camera (40 meters for all the methods), and calculating the standard metric of Intersection over Union (IoU) based on labeled ground truth left images. IoU is defined as TP/(TP+FN+FP), where T/F P/N stands for true/false positive/negative.

Yang et al.~\cite{yang2017semantic} \emph{exclude} the data that has not been projected onto images (gray color in the projected images), \emph{even when there exists corresponding ground truth data of it} (as shown in the ground truth images in Fig.~\ref{fig:kitti_05_proj_first}). For a fair comparison, we follow this approach for all methods and call it \emph{IoU Exclusive}. However, this evaluation ignores the classification error of gaps in the map, and cannot show the advantage of continuous mapping. Therefore, we compute a more rigorous \emph{IoU} by taking all projected data except the sky class into account.

\begin{table}
\centering
\caption{Quantitative results on KITTI odometry sequence 05 test set~\cite{kundu2014joint} for 8 common semantic classes, containing 40 images.}
\label{table:kitti05}
\resizebox{\columnwidth}{!}{
\begin{tabular}{llccccccccc}
{\bf Metric} & \multicolumn{1}{l}{\bf Method}
& \cellcolor{buildingColor}\rotatebox{90}{\color{white}Building} &
\cellcolor{roadColor}\rotatebox{90}{\color{white}Road} &
\cellcolor{vegetationColor}\rotatebox{90}{\color{white}Vege.} &
\cellcolor{sidewalkColor}\rotatebox{90}{\color{white}Sidewalk} &
\cellcolor{carColor}\rotatebox{90}{\color{white}Car} & 
\cellcolor{signateColor}\rotatebox{90}{\color{white}Signate} & 
\cellcolor{fenceColor}\rotatebox{90}{\color{white}Fence} &
\cellcolor{poleColor}\rotatebox{90}{\color{white}Pole} & \rotatebox{90}{\bf Average} \\ \hline 

\vspace{-2mm} \\

\multirow{3}{*}{IoU Exclusive}
& Yang et al.~\cite{yang2017semantic} & 86.2 & 91.5 & \bf 85.3 &  74.1 & \bf 77.1 & 16.8 & 78.5 & 28.0 & 67.2 \\
& BGKOctoMap-CRF & 86.1 & 88.0 & 82.3 & 73.6 & 71.9 & 15.5 & 73.8 & 27.7 & 64.9 \\
& S-CSM & 86.3 & 93.2 & 84.3 & \bf 80.0 & 76.8 & \bf 25.5 & 77.5 & \bf 30.1 & \bf 69.2  \\
& S-BKI & \bf 87.4 & \bf 93.3 & 84.7 & 79.9 & 76.9 & 18.6 & \bf 78.7 & 29.2 & 68.6
\\
\midrule

\multirow{3}{*}{IoU}
& Yang et al.~\cite{yang2017semantic} & 32.5 & 70.1 & 45.2 &  55.7 & 39.5 & 13.0 & 46.6 & 18.9 & 40.2\\
& BGKOctoMap-CRF & 43.5 & 70.9 & 49.4 & 55.5 & 40.2 & 12.7 & 46.4 & 13.9 & 41.6 \\
& S-CSM & 40.2 & 74.1 & 49.5 & 62.1 & 42.1  & \bf 20.3 & 47.7 & 22.8 & 44.9 \\
& S-BKI & \bf 45.6 & \bf 75.5 & \bf 52.8 & \bf 62.9 & \bf 43.3 & 14.9 & \bf 49.3 & \bf 22.9 & \bf 46.0
\\ \bottomrule
\end{tabular}
}
\end{table}

\begin{table}
\centering
\caption{Quantitative results on KITTI odometry sequence 15 test set~\cite{sengupta2013urban} for 8 common semantic classes, containing 25 images.}
\label{table:kitti15}
\resizebox{\columnwidth}{!}{
\begin{tabular}{llccccccccc}
{\bf Metric} & \multicolumn{1}{l}{\bf Method}
& \cellcolor{buildingColor}\rotatebox{90}{\color{white}Building} &
\cellcolor{roadColor}\rotatebox{90}{\color{white}Road} &
\cellcolor{vegetationColor}\rotatebox{90}{\color{white}Vege.} &
\cellcolor{sidewalkColor}\rotatebox{90}{\color{white}Sidewalk} &
\cellcolor{carColor}\rotatebox{90}{\color{white}Car} & 
\cellcolor{signateColor}\rotatebox{90}{\color{white}Signate} & 
\cellcolor{fenceColor}\rotatebox{90}{\color{white}Fence} &
\cellcolor{poleColor}\rotatebox{90}{\color{white}Pole} & \rotatebox{90}{\bf Average} \\ \hline 

\vspace{-2mm} \\

\multirow{3}{*}{IoU Exclusive}
& Yang et al.~\cite{yang2017semantic} & \bf 95.6 & 90.4 & \bf 92.8 & 70.0 & 94.4 & 0.1 & \bf 84.5 & 49.5 & 72.2 \\
& BGKOctoMap-CRF & 94.7 & 93.8 & 90.2 & 81.1 & 92.9 & 0.0 & 78.0 & 49.7 & 72.5 \\
& S-CSM & 94.4 & \bf 95.4 & 90.7 & \bf 84.5 & 95.0  & 22.2 & 79.3 & \bf 51.6 &  76.6 \\
& S-BKI & 94.6 & \bf 95.4 & 90.4 & 84.2 & \bf 95.1 & \bf 27.1 & 79.3 & 51.3 & \bf77.2
\\
\midrule

\multirow{3}{*}{IoU}
& Yang et al.~\cite{yang2017semantic} & 32.9 & 85.8 & 59.0 & \bf 79.3 & 61.0 & 0.9 & 46.8 & 33.9 & 50.0 \\
& BGKOctoMap-CRF & \bf 50.0 & 86.6 & 64.1 & 74.9 & 61.0 & 0.0 & 47.5 & \bf 36.7 & 52.6  \\
& S-CSM & 42.6 & 87.3 & 62.9 & 77.9 & 62.6 & 17.1 & 47.7 & 34.8 & 54.1 \\
& S-BKI & 49.3 & \bf 88.8 & \bf 69.1 & 78.2 & \bf 63.6 & \bf 22.0 & \bf 49.3 & \bf 36.7 & \bf 57.1 \\
\bottomrule
\end{tabular}
}
\end{table}

The quantitative results are given in Table~\ref{table:kitti05} and~\ref{table:kitti15}, where the two metrics are computed. For this experiment, the average runtime of Yang et al. is 4.41 $\sec$/scan, BGKOctoMap-CRF is 1.10 $\sec$/scan, S-CSM is 0.75 $\sec$/scan, and S-BKI is 0.36 $\sec$/scan. S-BKI has the highest IoU among almost all semantic classes compared with other maps, and S-CSM is the second-best method. We reiterate that the IoU Exclusive is not a reasonable metric for mapping performance evaluations; nevertheless, S-CSM and S-BKI still outperform the compared baselines using this metric. In the latter case, as expected, S-CSM and S-BKI perform similarly.

BGKOctoMap-CRF has a higher IoU than Yang's method because of the continuous occupancy model of BGKOctoMap. The gaps in the measurements are interpolated and CRF fills the labels from adjacent voxels. S-CSM outperforms both CRF-based methods, because even if the 3D CRF model further optimizes the grid labels, it is only post-processing pre-calculated occupied grids and, therefore, it cannot recover the correct semantic labels for misclassified occupancy or unknown grids. Specifically, even if BGKOctoMap-CRF is a continuous model for occupancy, it is not continuous for semantics and color. Thus, the predicted occupied voxels might not contain observation of semantics and color, leading to improper initialization of them for CRF potentials. In contrast, the counting sensor model uses a statistical model to infer the grid statistics. By adding the Bayesian kernel inference, S-BKI outperforms S-CSM as it can fill the gaps in the map using nearby measurements. Even for fully observed regions, by considering local correlations the map becomes more robust to noisy measurement.

\begin{table*}[t]
\centering
\caption{Mean IoU on SemanticKITTI dataset sequence 00-21~\cite{behley2019iccv} for 19 semantic classes. SqueezeSegV2-kNN (Sq.-kNN). Darknet53-kNN (Da.-kNN). Training (00-07, 09-10), Validation (8), Test (11-21).}
\label{table:semantickitti}
\resizebox{\textwidth}{!}{
\begin{tabular}{llllccccccccccccccccccc}
{\bf Seq.} & 
\multicolumn{1}{l}{\bf Method}&
\cellcolor{scarColor}\rotatebox{90}{\color{white}Car} &
\cellcolor{sbicycleColor}\rotatebox{90}{\color{white}Bicycle} &
\cellcolor{smotorcycleColor}\rotatebox{90}{\color{white}Motorcycle} &
\cellcolor{struckColor}\rotatebox{90}{\color{white}Truck} & 
\cellcolor{sothervehicleColor}\rotatebox{90}{\color{white}Other Vehicle} & 
\cellcolor{spersonColor}\rotatebox{90}{\color{white}Person} &
\cellcolor{sbicyclistColor}\rotatebox{90}{\color{white}Bicyclist} &
\cellcolor{smotorcyclistColor}\rotatebox{90}{\color{white}Motorcyclist} &
\cellcolor{sroadColor}\rotatebox{90}{\color{white}Road} &
\cellcolor{sparkingColor}\rotatebox{90}{\color{white}Parking} &
\cellcolor{ssidewalkColor}\rotatebox{90}{\color{white}Sidewalk} &
\cellcolor{sothergroundColor}\rotatebox{90}{\color{white}Other Ground} &
\cellcolor{sbuildingColor}\rotatebox{90}{\color{white}Building} &
\cellcolor{sfenceColor}\rotatebox{90}{\color{white}Fence} &
\cellcolor{svegetationColor}\rotatebox{90}{\color{white}Vegetation} &
\cellcolor{strunkColor}\rotatebox{90}{\color{white}Trunk} &
\cellcolor{sterrainColor}\rotatebox{90}{\color{white}Terrain} &
\cellcolor{spoleColor}\rotatebox{90}{\color{white}Pole} &
\cellcolor{strafficsignColor}\rotatebox{90}{\color{white}Traffic Sign} &

\rotatebox{90}{\bf Average}\\ \hline 

\vspace{-2mm} \\
\multirow{6}{*}{Training}& 
Sq.-kNN & 88.2 & 14.4 & 45.7 & 67.3 & 60.9 & 33.3 & 58.7 & 63.1 & 92.6 & 62.0 & 81.3 & 49.2 & 77.3 & 63.6 & 76.7 & 34.5 & 71.5 & 32.8 & 49.5 & 59.1 \\
& S-CSM (w/ Sq.-kNN) & 92.6 & 21.6 & 62.2 & 73.1 & 70.6 & 44.1 & 80.3 & 67.4 & 94.3 & 70.9 & 85.1 & 52.3 & 82.0 & 69.1 & 81.4 & 47.8 & 75.4 & 50.8 & 65.0 & 67.7 \\
& S-BKI (w/ Sq.-kNN) & 93.5 & 29.1 & 73.9 & 82.0 & 77.0 & 54.6 & 87.2 & 73.7 & 93.8 & 73.6 & 84.2 & 55.7 & 83.8 & 70.1 & 82.8 & 53.9 & 75.9 & 54.6 & 70.4 & 72.1 \\
& Da.-kNN & 94.7 & 42.3 & 81.8 & 83.4 & 69.4 & 69.4 & 72.5 & 53.7 & 96.7 & 88.6 & 92.6 & 82.1 & 95.4 & 85.0 & 92.0 & 71.0 & 88.3 & 70.6 & 82.4 & 79.6 \\     
& S-CSM (w/ Da.-kNN) & 96.0 & 48.6 & 88.3 & 84.5 & 71.4 & 77.3 & 83.6 & 54.3 & \bf 96.8 & 89.7 & \bf 93.3 & 84.2 & 96.6 & 86.7 & 93.5 & 79.2 & 90.0 & 80.0 & 88.9 & 83.3 \\
& S-BKI (w/ Da.-kNN) & \bf 96.9 & \bf 53.2 & \bf 90.9 & \bf 85.9 & \bf 73.3 & \bf 83.5 & \bf 88.4 & \bf 59.8 & \bf 96.8 & \bf 89.9 & 93.1 & \bf 85.4 & \bf 97.3 & \bf 87.4 & \bf 94.2 & \bf 81.3 & \bf 90.9 & \bf 82.0 & \bf 90.6 & \bf 85.3 \\
\midrule
\multirow{6}{*}{Validation}& 
Sq.-kNN & 86.7 & 14.4 & 24.6 & 21.0 & 23.3 & 23.5 & 40.9 & 0.0 & 90.1 & 32.4 & 74.8 & \bf 1.2 & 79.6 & 42.7 & 79.2 & 36.5 & 71.1 & 28.3 & 24.8 & 41.8 \\
& S-CSM (w/ Sq.-kNN) & 90.5 & 23.0 & 34.9 & 26.8 & 29.1 & 32.4 & 49.4 & 0.0 & 92.6 & 38.7 & 79.0 & 1.1 & 84.6 & 51.6 & 83.3 & 48.3 & 72.9 & 44.1 & 31.6 & 48.1 \\
& S-BKI (w/ Sq.-kNN) & 92.3 & 30.0 & 39.7 & 29.3 & \bf 32.1 & 38.8 & 54.7 & 0.0 & 92.9 & 40.9 & 79.9 & 1.1 & 86.6 & 54.6 & 84.9 & 52.3 & 74.2 & 47.9 & 34.7 & 50.9 \\
& Da.-kNN & 91.0 & 25.0 & 47.1 & 40.7 & 25.5 & 45.2 & 62.9 & 0.0 & 93.8 & 46.5 & 81.9 & 0.2 & 85.8 & 54.2 & 84.2 & 52.9 & 72.7 & 53.2 & 40.0 & 52.8\\
& S-CSM (w/ Da.-kNN) & 92.6 & 32.5 & 54.9 & 43.4 & 26.2 & 51.3 & 69.2 & 0.0 & \bf 94.6 & 49.2 & \bf 84.0 & 0.1 & 87.9 & 58.4 & 85.8 & 59.9 & 73.3 & 61.7 & 43.0 & 56.2 \\
& S-BKI (w/ Da.-kNN) & \bf 93.5 & \bf 33.5 & \bf 57.3 & \bf 44.5 & 27.2 & \bf 52.9 & \bf 72.1 & 0.0 & 94.4 & \bf 49.6 & \bf 84.0 & 0.0 & \bf 88.7 & \bf 59.6 & \bf 86.9 & \bf 62.5 & \bf 75.3 & \bf 63.6 & \bf 45.1 & \bf 57.4 \\
\midrule
\multirow{2}{*}{Test}& 
Da.-kNN & 82.4 & 26.0 & 34.6 & 21.6 & 18.3 & 6.7 & 2.7 & \bf 0.5 & 91.8 & 65.0 & 75.1 & 27.7 & 87.4 & 58.6 & 80.5 & 55.1 & 64.8 & 47.9 & 55.9 & 47.5 \\
& S-BKI (w/ Da.-kNN) & \bf 83.8 & \bf 30.6 & \bf 43.0 & \bf 26.0 & \bf 19.6 & \bf 8.5 & \bf 3.4 & 0.0 & \bf 92.6 & \bf 65.3 & \bf 77.4 & \bf 30.1 & \bf 89.7 & \bf 63.7 & \bf 83.4 & \bf 64.3 & \bf 67.4 & \bf 58.6 & \bf 67.1 & \bf 51.3\\
\bottomrule
\end{tabular}
}
\end{table*}

\begin{table}
\centering
\caption{Comparison of map quality using the Area Under ROC Curve (AUC) and runtime of the four methods on the example shown in Fig.~\ref{fig:stanford_semantics}.}
\label{table:occupancy}
\begin{tabular}{lllll}
\toprule
Method & OctoMap & BKIOctoMap & S-CSM & S-BKI \\
 \midrule
AUC & 0.7226 & \bf 0.7801 & 0.7274 & \bf 0.7801\\ 
 \midrule
Runtime (s) & 252.32 & 73.30 & 480.44 & \bf 68.17 \\ 
 \bottomrule
\end{tabular}
\end{table}

\subsection{SemanticKITTI Dataset}
SemanticKITTI~\cite{behley2019iccv} is a large-scale dataset based on the KITTI odometry dataset. It provides dense annotations for each LiDAR scan of 22 sequences including camera poses estimated from a surfel-based SLAM approach (SuMa)~\cite{behley2018rss}. The input data of this dataset is collected by a Velodyne HDL-64E laser scanner. The semantic measurements are generated by RangeNet++~\cite{milioto2019iros}, which is a state-of-the-art LiDAR-only semantic segmentation deep neural network. To investigate mapping performance on noisy data, we choose two backends provided in RangeNet++: the best-performing one, Darknet53-kNN, and SqueezeSegV2-kNN which has lower performance. All maps are built at a resolution of 0.1 $\mathrm{m}$ and without any pre-processing of the input data. 

For evaluation, we use all sequences in SemanticKITTI. For training (00-07, 09-10) and validation (08) sequences, we compare the 3D predictions with ground truth labels. To obtain the IoU metrics for test (11-21) sequences, we submitted our results to the official evaluation server which are shown on the multi-scan leaderboard\footnote{\href{https://competitions.codalab.org/competitions/20331\#results}{https://competitions.codalab.org/competitions/20331\#results} (ganlumm)}. As our method is for static environments, we cannot differentiate between static and dynamic objects. For static semantic classes, we outperforms Darknet53-kNN for 18 out of 19 classes on test sequences.

Quantitative results on all sequences are given in Table~\ref{table:semantickitti}. For this experiment, the average runtime of S-CSM is 9.48 $\sec$/scan, S-BKI is 1.67 $\sec$/scan. For all sequences, our semantic mapping methods can improve the prior segmentation IoU by fusing multiple scans. We note that S-BKI consistently outperforms S-CSM in almost all semantic classes, which shows the advantage of Bayesian kernel inference and continuous semantic maps. When S-CSM does outperform S-BKI, the IoUs are close to each other. Moreover, the mapping improvement over SqueezeSegV2-kNN is much higher than Darknet53-kNN, which shows our methods can deal with noisy input data.

\begin{figure}[t]
\centering 
  \subfloat{\includegraphics[width=.99\columnwidth,trim={0cm 0cm 0cm 0cm},clip]{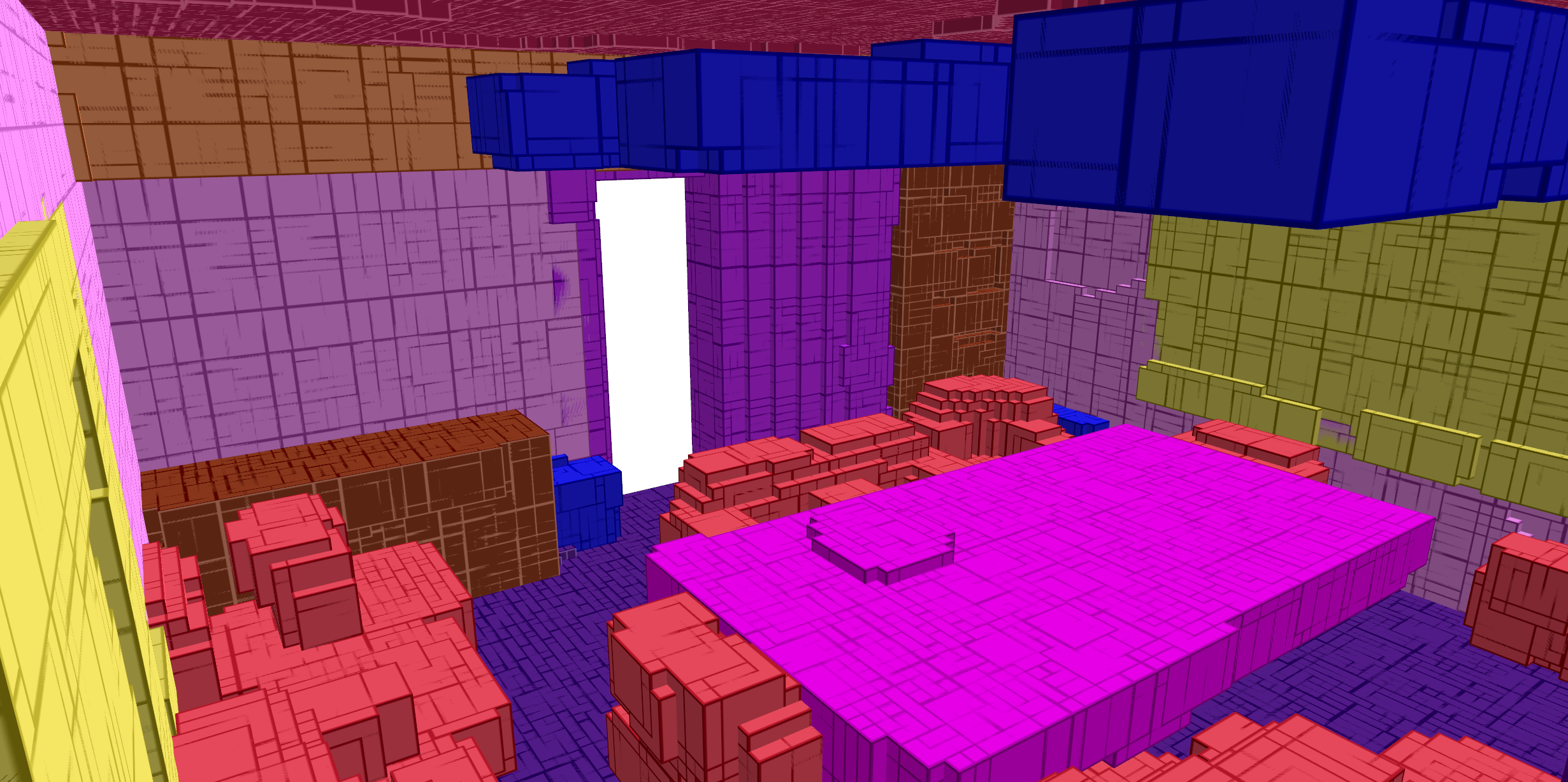}}\\ \squeezeup
\subfloat{\crule[beamColor]{0.3cm}{0.2cm} \scriptsize{Beam}
\crule[boardColor]{0.3cm}{0.2cm} \scriptsize{Board}
\crule[bookcaseColor]{0.3cm}{0.2cm} \scriptsize{Bookcase}
\crule[ceilingColor]{0.3cm}{0.2cm} \scriptsize{Ceiling}
\crule[chairColor]{0.3cm}{0.2cm} \scriptsize{Chair}} \\\squeezeup
\subfloat{\crule[clutterColor]{0.3cm}{0.2cm} \scriptsize{Clutter}
\crule[doorColor]{0.3cm}{0.2cm} \scriptsize{Door}
\crule[floorColor]{0.3cm}{0.2cm} \scriptsize{Floor}
\crule[tableColor]{0.3cm}{0.2cm} \scriptsize{Table}
\crule[wallColor]{0.3cm}{0.2cm} \scriptsize{Wall}} \\
\caption{S-BKI map of a conference room in Area 3 of Stanford 2D-3D-Semantics Dataset~\cite{2017arXiv170201105A}.}
\label{fig:stanford_semantics}
\end{figure}

\subsection{Occupancy Evaluation}
To support the claim that S-BKI is a semantic occupancy mapping method, we evaluate the accuracy of occupancy prediction of S-CSM, S-BKI, OctoMap and BGKOctoMap. The experiment is performed using a conference room in Area 3 of Stanford 2D-3D-Semantics Dataset~\cite{2017arXiv170201105A}, as ground-truth occupancy values are provided. For S-CSM and S-BKI, we use the annotated point clouds to build the semantic occupancy maps, and the same point clouds without semantics for OctoMap and BGKOctoMap. The semantic map built by S-BKI is shown in Fig.~\ref{fig:stanford_semantics}. Comparisons of map quality and runtime of the four methods are given in Table~\ref{table:occupancy}. For S-CSM and S-BKI, the probability of occupancy is computed as the sum of all probabilities of valid semantic classes. Among all methods, S-BKI and BGKOctoMap have the highest (identical) performance, which shows that S-BKI reduces to BGKOctoMap when only occupancy is of interest, not only theoretically, but also experimentally. S-CSM is slower than S-BKI because the block depth is set to one, thus S-CSM has more blocks to be computed.

\subsection{Impact of Parameters}
\begin{figure}[t]
    \centering
    \subfloat{\includegraphics[width=0.5\columnwidth, trim={0cm 0cm 1cm 0.25cm},clip]{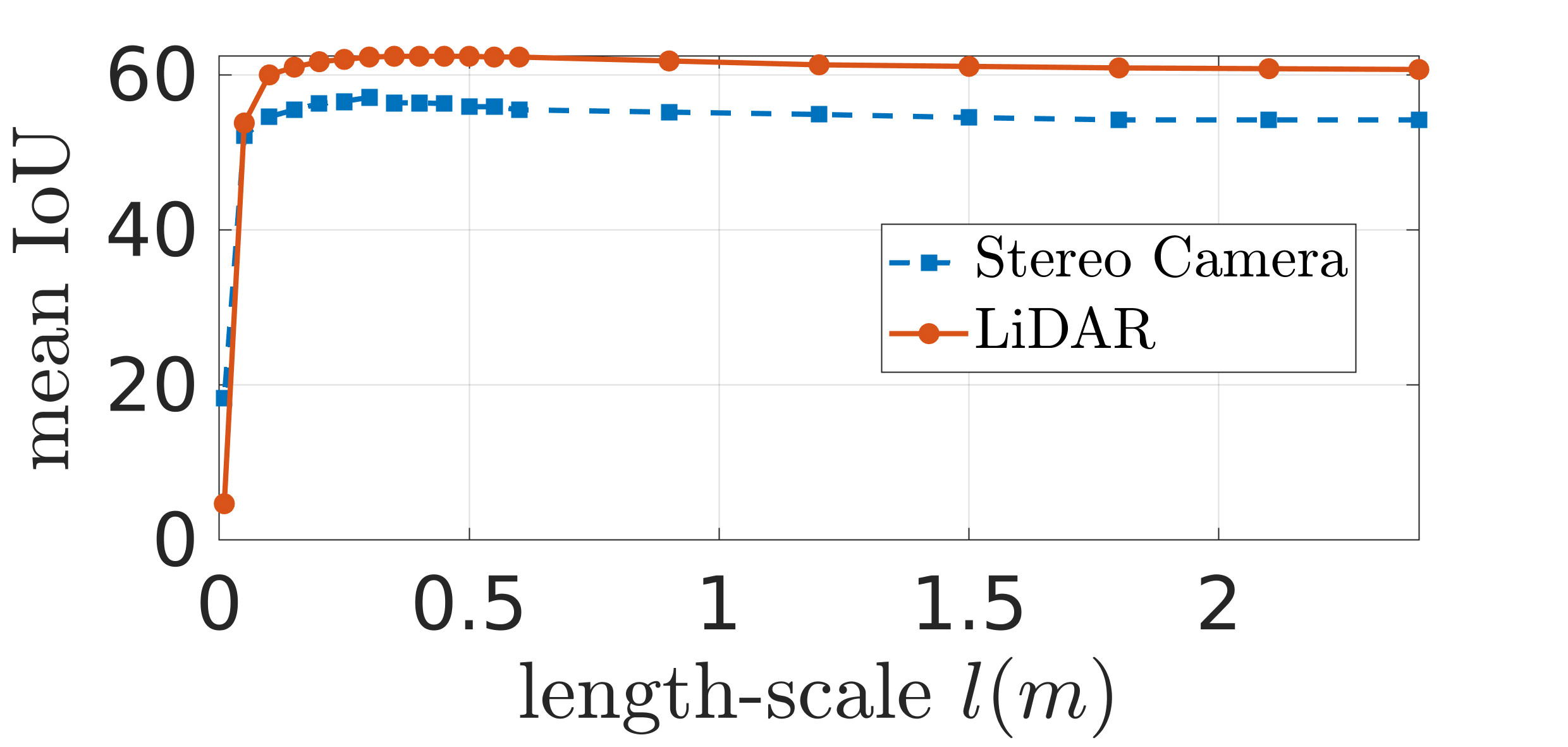}}
    \subfloat{\includegraphics[width=0.5\columnwidth, trim={0cm 0cm 1cm 0.25cm},clip]{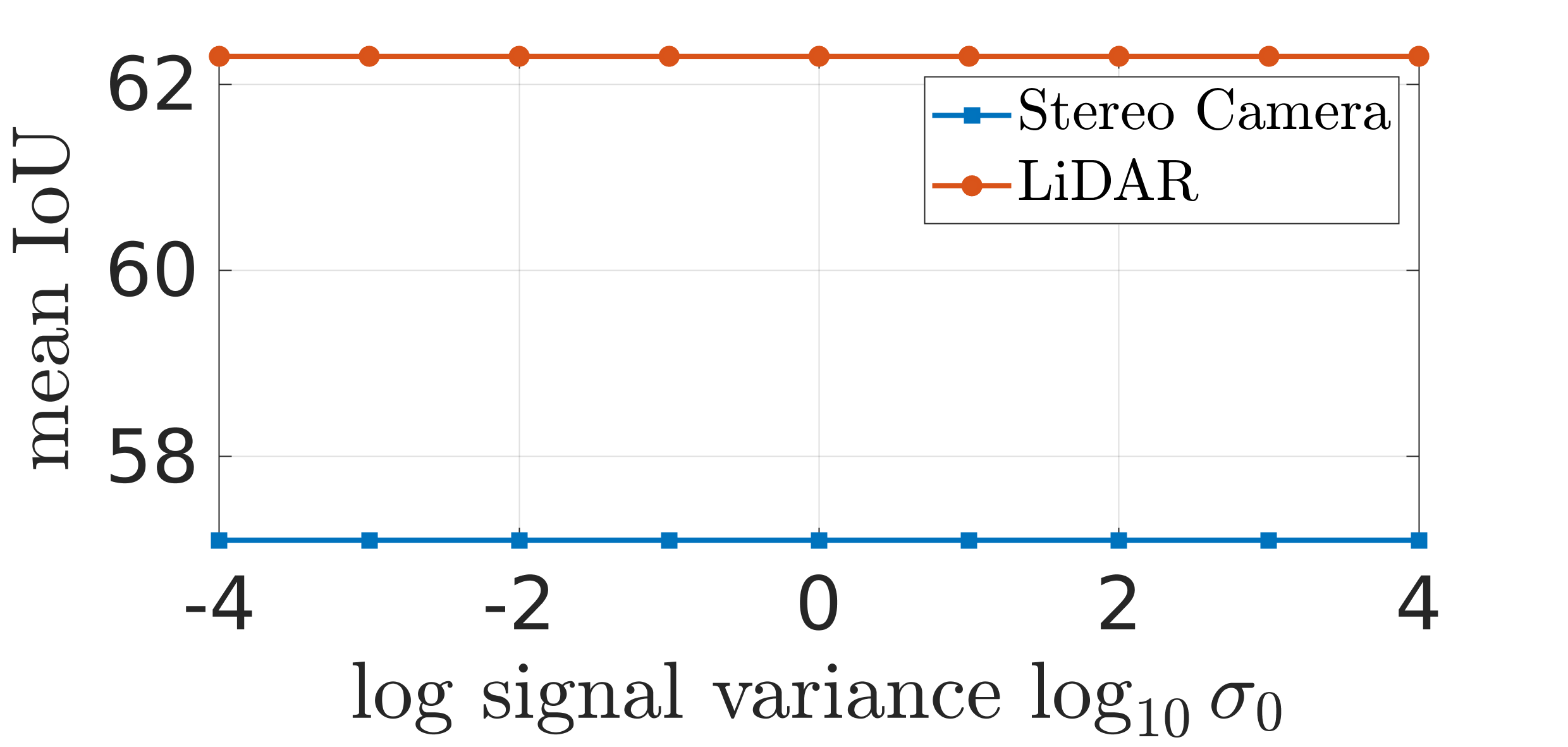}}
    \caption{Impact of parameters on mapping performance for KITTI dataset sequence 15~\cite{sengupta2013urban} (stereo camera) and SemanticKITTI dataset sequence 04~\cite{behley2019iccv} (LiDAR). Only one parameter at a time is varied while the others are kept at the values in Table~\ref{table:parameters}. Both figures show reasonable robustness to the parameter variations.}
    \label{fig:impact}
\end{figure}

We empirically study the sensitivity of S-BKI mapping to the kernel length-scale and signal variance. The experiments are conducted using KITTI dataset sequence 15~\cite{sengupta2013urban} for stereo camera and SemanticKITTI datase sequence 04~\cite{behley2019iccv} for LiDAR data. All other parameters are fixed to the values indicated in Table~\ref{table:parameters}. In Fig.~\ref{fig:impact}, we plot the kernel length-scale $l$ and signal variance $\sigma_0$ against the mean IoU metrics. The influence of the kernel length-scale on mapping performance for both stereo camera and LiDAR data is similar: the mean IoU increases rapidly as the length-scale varies from 0.01 to 0.1, gradually increases to a peak value, and then drops gradually as the length-scale increases. S-BKI achieves the best performance when $l = 0.3$ for stereo camera data and $l = 0.4$ for LiDAR data. This is reasonable because LiDAR data is sparser than stereo camera data and longer distance should be considered. The mapping performance is insensitive to signal variance over a large scale; this is because we use the same signal variance for all semantic classes. To see an effect, one would need to allow signal variance to vary from class-to-class.

\subsection{Experimental Results on a Cassie Bipedal Robot}

\begin{figure}[t]
    \centering
    \includegraphics[width=0.48\columnwidth,trim={0 4cm 0 0},clip]{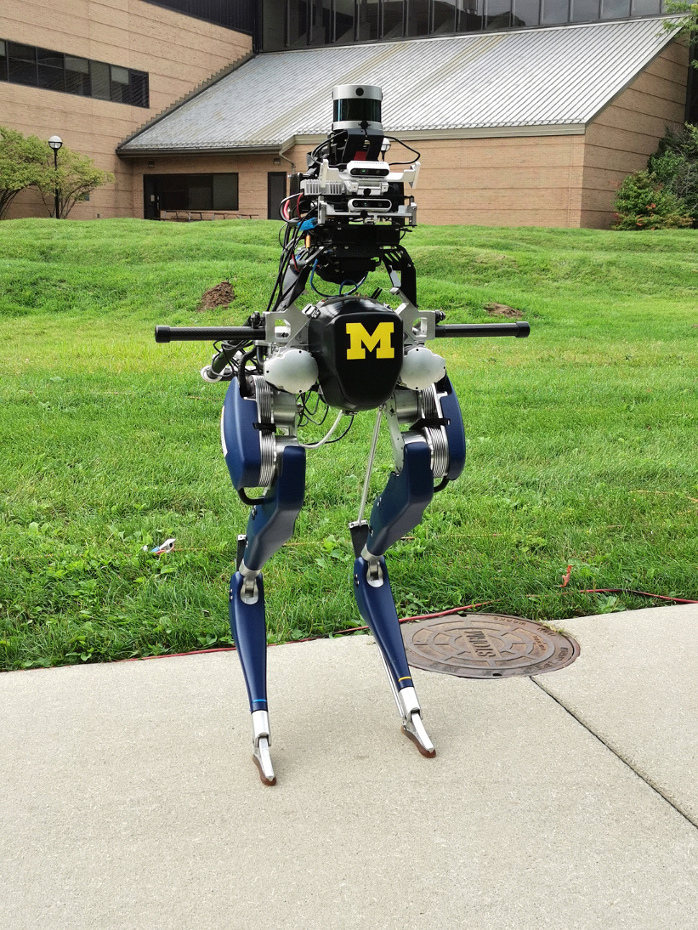} \hfill
    \includegraphics[width=0.49\columnwidth]{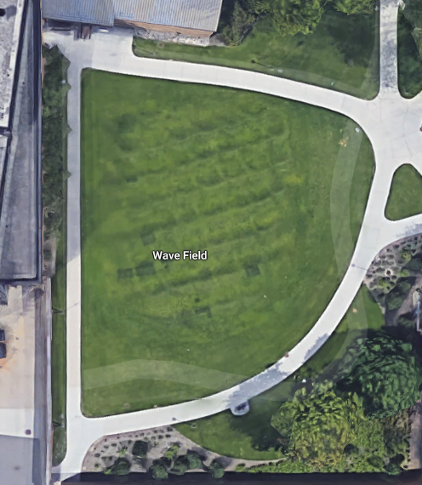}\\
    \includegraphics[width=0.45\columnwidth]{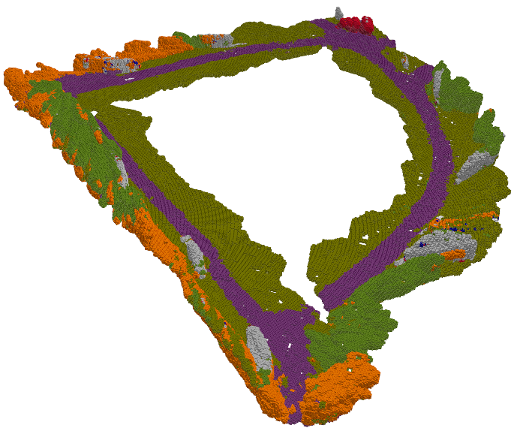} \hfill
    \includegraphics[width=0.4\columnwidth]{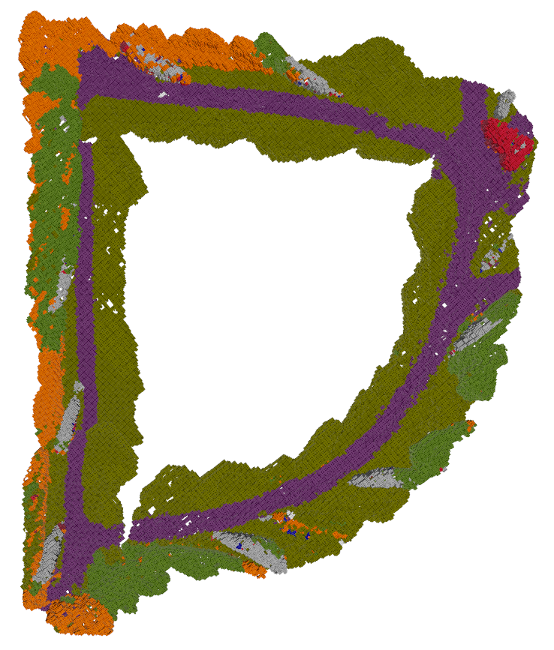}
    \caption{Top Left: Cassie Blue has a custom designed torso on which is mounted an Intel RealSense depth camera capable of providing both RGB images and corresponding organized point clouds in outdoor environments. Top right: Google satellite map of the Wave Field of the University of Michigan - North Campus. 
    Bottom: From left to right are the 3D and 2D views of S-BKI map. While the robot is navigating along the sidewalk, S-CSM produces discontinuous semantic maps from sparse sensor measurements, which may cause the robot's planner to regard the gaps in the map as unwalkable areas, a practical problem when we conduct autonomous walking experiments with Cassie Blue (a video of the experiment is available at \href{https://www.youtube.com/watch?v=uFyT8zCg1Kk\&t=3s}{https://www.youtube.com/watch?v=uFyT8zCg1Kk\&t=3s}). S-BKI model produces a continuous and smooth map, where gaps are assigned with labels inferred from local correlations in the map.}
    \label{fig:cassie_pic}
\end{figure}

We test our mapping methods on data collected using the bipedal robot Cassie Blue shown in Fig~\ref{fig:cassie_pic}. To obtain semantic measurements, we manually annotated 1194 training images and 457 validation images from the NCLT dataset~\cite{carlevaris2016university}. The NCLT dataset was selected because it shares a similar environmental domain as the Wave Field data, which includes \emph{background, water, road, sidewalk, terrain, building, vegetation, car, person, bike, pole, stair, traffic sign and sky} for a total of 14 classes. We used these images to fine-tune a modified 2D segmentation network MobileNet~\cite{siam2018rtseg} with a pre-trained model on the ImageNet dataset~\cite{deng2009imagenet} for efficiency. The fine-tuned network segments the RGB images, and then we can directly label the organized point clouds. 

The qualitative results are given in Fig.~\ref{fig:cassie_pic}. To show the mapping performance of our methods on sparse data, we downsample the point clouds per scan to a resolution of 0.2 $\mathrm{m}$, and build a semantic occupancy map with a resolution of 0.1 $\mathrm{m}$. S-BKI runs at about 2 $\mathrm{Hz}$. The mapping drift after one full round of the Wave Field is because of the odometry system~\cite{Hartley-RSS-18} instead of SLAM used in the experiment. 

\section{Discussions and Limitations}
\label{sec:discussions}
In practice, semantic measurements do not necessarily come in the form of a one-hot vector, but rather a pseudo-probability vector obtained from the softmax output of a classifier. Taking the max rather than the softmax results in the current formulation. Taking the softmax, on the other hand, results in other models corresponding to a set of model-averaging techniques (i.e., the linear opinion pooling and Nadaraya-Watson kernel-regression) that are similar, but not identical, to the Bayesian model presented in Sec.~\ref{sec:csm}.

There are still several limitations to this work. First, the length-scale of the kernel function trades off predictive ability and classification accuracy. When the length-scale is large, the model can extrapolate large-scale trends in data, and thus be more predictive; however, the classification accuracy may drop for small objects in the environment. In the current approach, we manually tune the length-scale and use the same scale everywhere, independent of the class. Optimizing the hyperparameters in a Bayesian framework can be helpful. In addition, varying the length-scale and signal variance based on geometric features and semantic properties may further improve semantic mapping performance. Secondly, the memory and space storage for large-scale mapping is another limitation. We currently store the entire semantic map in computer memory without any pruning. However, with the current test-data octrees data structure, even when storing the map after pruning, the save in memory consumption is not substantial. How to compress the continuous semantic maps is an interesting future research direction. Finally, the current semantic map is for static environments, differentiating between static and dynamic semantic labels is also an interesting future work.

\section{Conclusion}
\label{sec:conclusion}

In this paper, we extended the counting sensor model for occupancy grid mapping to a semantic counting sensor model for semantic occupancy mapping. To relax the independent-grid assumption in occupancy grid mapping, we used a Bayesian spatial kernel inference to generalize the semantic counting sensor model to continuous semantic mapping. Extensive experimental results show the proposed methods work with both dense stereo camera and LiDAR data. We improved the mapping performance over the state-of-the-art semantic mapping system using the KITTI dataset, and increased the segmentation accuracy over a 3D deep neural network with kNN processing using the SemanticKITTI dataset. We labeled the NCLT dataset and collected data using Cassie Blue biped robot to further evaluate the mapping performance in real world experiments. The S-BKI model consistently outperforms S-CSM, which shows the advantage of using Bayesian kernel inference in continuous mapping.


\section*{Acknowledgment}

This article solely reflects the opinions and conclusions of its authors and not TRI or any other Toyota entity. The authors would like to thank Yukai Gong for the development of the feedback controller utilized in the Cassie experiments as well as Bruce Huang, Zhenyu Gan, Omar Harib, Eva Mungai, and Grant Gibson for their help in collecting experimental data.

\bibliographystyle{IEEEtran}
\bibliography{strings-abrv,ieee-abrv,refs}

\end{document}